\documentclass{article}
\usepackage{colt09e,times}
\usepackage{amsmath,amsfonts,amssymb}
\usepackage{fancyhdr}
\usepackage{lastpage}
\usepackage{extramarks}
\usepackage{chngpage}
\usepackage{soul}
\usepackage[usenames,dvipsnames]{color}
\usepackage{graphicx,float,wrapfig}
\usepackage{ifthen}
\usepackage{listings}
\usepackage{courier}
\usepackage{enumerate}
\usepackage{ctable,url}

\begin{document}

\title{Survey \& Experiment: Towards the Learning Accuracy}

\author{Zeyuan Allen Zhu \\ Tsinghua University\\ {\tt zhuzeyuan@hotmail.com}}

\maketitle

\begin{abstract}

To attain the best learning accuracy, people move on with difficulties and frustrations.
Though one can optimize the empirical objective using a given set of samples,
 its generalization ability to the entire sample distribution remains questionable.
Even if a fair generalization guarantee is offered, one still wants to know what is to happen if the regularizer is removed,
 and/or how well the artificial loss (like the hinge loss) relates to the accuracy.

For such reason, this report surveys four different trials towards the learning accuracy, embracing the major advances in supervised learning theory in the past four years.
Starting from the generic setting of learning, the first two trials introduce the best optimization and generalization bounds for convex learning, and the third trial gets rid of the regularizer.
As an innovative attempt, the fourth trial studies the optimization when the objective is exactly the accuracy, in the special case of binary classification. This report also analyzes the last trial through experiments.

\end{abstract}

\newcommand{\vw}{\mathbf{w}}
\newcommand{\vv}{\mathbf{v}}
\newcommand{\vx}{\mathbf{x}}
\newcommand{\vt}{\boldsymbol{\theta}}

\section{Introduction}
A generic learning problem can be regarded as an optimization over parameter $\vw \in \mathcal{W}$,
 and the objective function is given by $f(\vw;\vt)$ where $\vt$ is a given sample.
An empirical objective $\hat{F}(\vw)$ can be written as
$$ \hat{F}(\vw)=\hat{\mathbb{E}}[f(\vw; \vt)]=\frac{1}{m}\sum_{i=1}^m f(\vw;\vt_i)\enspace,$$
here $\vt_1,\vt_2,\ldots,\vt_m$ are a sequence of observed samples.
It is normally assumed that these samples are i.i.d. drawn from some unknown distribution $D$,
 and therefore the stochastic objective $F(\vw)$ is often more desirable:
$$ F(\vw)=\mathbb{E}_{\vt \sim D}[f(\vw;\vt)] \enspace.$$

For example, if we take $f(\vw; \vt=(\vx,y)) = \max\{0,1-y\langle \vw,\vx \rangle\} + \frac{\lambda}{2}\|\vw\|_2^2$ we will arrive at the famous SVM, with a weighted $L_2$ norm regularizer.
The kernel trick can also be adopted which allows a non-linear prediction: $f(\vw; \vt) = \max\{0,1-y\langle \vw,\phi(\vx) \rangle\} + \frac{\lambda}{2}\|\vw\|_2^2$.

\textbf{Trial 1.}To obtain a good learning accuracy, the first trial is to optimize over the empirical objective $\hat{F}(\vw)$.
We normally assume that $f(\cdot;\vt)$ is a convex function because we can then utilize optimization techniques like interior point method or gradient descent to conquer the minimization.
Recent results in for example \cite{Mirror,PEGASOS} have shown that in many cases, using the stochastic gradient descent experiences the fastest runtime in minimizing the empirical objective, and \cite{P-pack} has generalized this in the extent of using kernels and (batched) parallel computation.
In Section \ref{sec:step1}, we will briefly describe the main framework of \cite{Mirror}, because it embraces all previously known first-order algorithms as special cases.

\textbf{Trial 2.} To better understand the learning accuracy to future samples, we need to establish the connection between the stochastic objective $F(\vw)$ and empirical objective $\hat{F}(\vw)$. This is called the {\em generalization}.
A recent paper \cite{stochastic} completed a thorough classification over the types of learning problems, and tells us how well each of them guarantees a good generalization error bound. The main results of this paper is provided in Section \ref{sec:step2} as a good reference, but this is only the second trial.

As a partial summary, in the above two trials, when $f$ is convex both results have an error bound proportional to $\frac{1}{\sqrt{T}}$.
In Step 1, this $T$ is the number of iterations (which is also proportional to the runtime) and the error bound is the difference between the optimal solution and the one generated by the SGD.
In Step 2, this $T$ is the number of samples and the error bound is (imprecisely) the difference over the stochastic and empirical objective.
However, if we further require $f$ to be strongly-convex, by for instance adding a regularizer, both bounds immediately decrease to $\frac{1}{T}$ instead of $\frac{1}{\sqrt{T}}$, and this partially explains why we add a regularizer from a theoretical point of view.

\textbf{Trial 3.} Now comes to the third trial, an attempt to bound the stochastic loss instead of the stochastic objective.
As explained above, we often add a regularizer (e.g. $r(\vw)=\frac{\lambda}{2}\|\vw\|^2$) to the objective function $f(\vw;\vt)=l(\vw;\vt)+r(\vw)$.
Therefore, even if we achieve a close-to-optimal solution for the stochastic objective $F(\vw)$, it is still far away from the stochastic loss
$$L(\vw)=\mathbb{E}_{\vt \sim D}[l(\vw;\vt)]\enspace.$$
To further build a connection between these two quantities, \cite{inverse, inverse2} used a so-called {\em oracle inequality} and deduced a final bound for this stochastic loss, with respect to the running time of a program, and the number of training samples $m$.
This work was recognized by the best paper awards of ICML 2008 and ICDM 2009, and briefly described in Section \ref{sec:step3}.

\textbf{Trial 4.} However, how well such loss function characterizes the word ``accuracy'' remains a problem.
One can feel free to use any convex loss functions (like hinge loss or logistic loss), but they do not reflect the accuracy at all.
In the paper of \cite{zeroone}, they consider the following {\em non-convex} objective:
$$f(\vw;\vt=(\vx,y))=\left| \frac{1}{2}\textrm{sgn}(\langle \vw, \phi(\vx)\rangle + \frac{1}{2} - y \right|$$
which is exactly the definition of accuracy (for binary classification) if $y \in \{0,1\}$.

Though incorporating the traditional Rademacher generalization bound \cite{Rademacher} one can still obtain good generalization guarantee,
 the empirical optimization becomes hard (Appendix A of \cite{zeroone}).
Realizing such difficulty, \cite{zeroone} studied {\em improper} learning and constructed a larger class of classifier.
Not only the empirical optimization in the new concept class is convex, the minimizer is also good enough so that original zero-one objective problem is learnable using such minimizer.
This paper was recognized by the best paper award of COLT 2010, and this report is also going to analyze its practical value against public data sets in Section \ref{sec:step4}.

\subsection{Preliminary}
For the lack of space, the definitions of strongly-convex, Lipschitz continuity, dual norm and Bregman divergence are ignored in this report. The readers who are interested in the technical details may refer to (nearly) any of the references attached to this report and look into its preliminary section.

\section{Trial 1: Empirical Optimization}\label{sec:step1}

We denote the empirical objective for the $t$-th sample as $$f(\vw;\vt_t)=f_t(\vw)+r(\vw)\enspace,$$ in which $r$ is a convex regularization function,
 and $f_t$ is a convex loss function associated with example $t$.

In the online and batch learning setting, we have a sequence of $T$ samples.
At the beginning the the $t$-th round, the algorithm must make a prediction $\vw_t$ and then receive a function $f_t$.
The ultimate goal is to minimize the following {\em regularized regret}:
$$R(T) = \max_{\vw^* \in \mathcal{W}} \sum_{t=1}^T [f_t(\vw_t)+r(\vw_t)-f_t(\vw^*)-r(\vw^*)]$$
in which $\vw^*$ is the optimal empirical minimizer for this sequence of samples.

The paper \cite{Mirror} defined the following update sequence
\begin{equation}\label{eq:mirror_update}
\vw_{t+1}=\mathop{\textrm{argmin}}_{\vw\in\mathcal{W}} \eta \langle \partial f_t(\vw_t),\vw \rangle + B_{\psi}(\vw,\vw_t)+\eta r(\vw)\enspace.
\end{equation}
Here $\eta$ is a parameter that is to be tuned later, $\partial$ is the sub-gradient, and $\vw_1$ can be set to zero vector.
The Bregman divergence associated with $\psi$ is defined as
$$ B_{\psi}(\vw,\vv)=\psi(\vw)-\psi(\vv)-\langle \nabla\psi(\vv),\vw-\vv\rangle \enspace, $$
and we require $\psi$ to be $\alpha$-strongly convex w.r.t. a norm $\|\cdot\|$, and then $B_{\psi}(\vw,\vv) \geq \frac{\alpha}{2}\|\vw-\vv\|^2$ for this $\alpha$.
A very simple choice is to let $\psi(\vw)=\frac{1}{2}\|\vw\|^2$ and then $B_{\psi}(\vw,\vv)=\frac{1}{2}\|\vw-\vv\|^2$.

One of the main theorems in \cite{Mirror} proves that:
\begin{theorem}\label{thm:mirror1}
Let $\psi$ be $\alpha$-strongly convex w.r.t. norm $\|\cdot\|$.
Suppose $\mathcal{W}$ is compact OR the function $f_t$ is $G$-Lipschitz $\|\partial f_t\|_* \leq G$.\footnote{$\|\cdot\|_*$ is the dual norm of $\|\cdot\|$.}
Then setting $\eta=\sqrt{2\alpha B_{\psi}(\vw^*,\vw_1)} / (G\sqrt{T})$,
$$ R_{\psi}(T) \leq \sqrt{2T B_{\psi}(\vw^*,\vw_1)} G/\sqrt{\alpha} \enspace. $$
\end{theorem}
Notice that this result does not use any property of the regularized term $r(\vw)$, and it holds even if $r(\vw)=0$.

This square root (w.r.t. $T$) regret allows one to design a stochastic gradient descent algorithm with random sampling over the training data set. However, to obtain an optimization error of $\epsilon$, one usually needs to run $T=\Omega(1/\epsilon^2)$ number of iterations. This is not good enough.

When the objective function $f_t(\vw)+r(\vw)$ is strongly-convex over $\vw$, w.l.o.g. we can let $r$ be $\lambda$ strongly-convex and $f$ be only of the classical convexity. In this case if we replace the update sequence of Eq.~\ref{eq:mirror_update} and let $\eta$ vary for different $t$. Specifically, we let $\eta_t=\frac{1}{\lambda t}$.
With only little extra effort, one can prove the following logarithmic regret bound:

\begin{theorem}\label{thm:mirror2}
Let $r$ be $\lambda$-strongly convex w.r.t. $\psi$, and $\psi$ be $\alpha$-strongly convex w.r.t. norm $\|\cdot\|$.
If function $f_t$ is $G$-Lipschitz $\|\partial f_t\|_* \leq G$. Then:
$$ R_{\psi}(T) = O\left( \frac{G^2}{\lambda\alpha} \log T \right) \enspace. $$
\end{theorem}

Notice that if we set $\psi(\vw)=\frac{1}{2}\|\vw\|^2_2$ and we immediately arrive at the online counterpart of the famous \texttt{PEGASOS} algorithm, which is the currently best-known linear SVM classifier \cite{PEGASOS}, and arguably the best known kernel SVM classifier at least under the parallel setting \cite{P-pack}.

In sum, depending on the convexity of the objective function, one may adopt either the convex or the strongly-convex version of the mirror descent algorithm summarized in \cite{Mirror}, with a satisfiable theoretical bound on the regret. Regarding the off-line problem given a fixed set of training samples, as long as in each iteration a sample is uniformly chosen at random from this training set, a similar bound can be deduced just like \cite{PEGASOS,inverse,inverse2}, using Markov inequality.

\section{Trial 2: Stochastic Optimization}\label{sec:step2}

As advertised in the introduction, although a sub-optimal $\vw$ is deduced from minimizing the empirical objective $\hat{F}(\vw)$, we are more interested in its generalization ability to $F(\vw)$.
In this section we follow the convention: use $\hat{\vw}$ to denote the empirical minimizer, and $\vw^*$ the stochastic minimizer.

  \begin{figure*}[htbp]
    \centering
    \includegraphics[width=1.8\columnwidth]{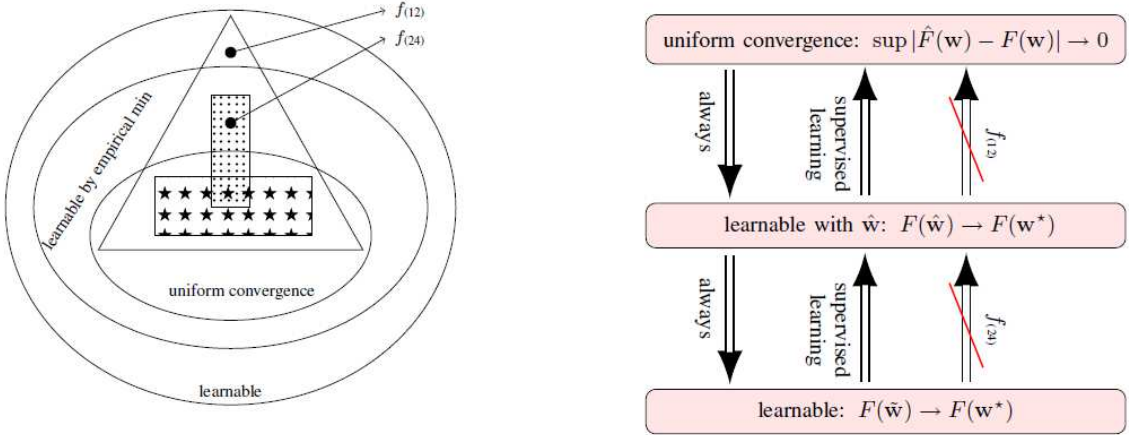}
    \caption{Three classes of learnable problems. Picture borrowed from \cite{stochastic}.}
    \label{stochastic.eps}
  \end{figure*}
The paper \cite{stochastic} analyzed three classes of problems (see Figure~\ref{stochastic.eps}). The first and the smallest class is the one that {\em guarantees uniform convergence}:
$$\sup_{\vw\in\mathcal{W}}\left| F(\vw)-\hat{F}(\vw)\right| \rightarrow 0\enspace,$$
which says that the difference between two kinds of objectives tends to zero as $m\rightarrow \infty$.
A relatively larger class is the one that is {\em learnable by empirical minimizer}:
$$ F(\hat{\vw})-F(\vw^*) \rightarrow 0\enspace,$$
which guarantees point-wise convergence at the empirical minimizer.
At last, the largest class of {\em learnable functions} is defined as
\begin{quote}
``there exists a rule for choosing $\tilde{\vw}$ based on samples, such that $F(\tilde{\vw})-F(\vw^*) \rightarrow 0$''.
\end{quote}

All of the {\em generalized linear problems}\footnote{Satisfying $f(\vw;\vt)=g(\langle \vw, \phi(\vt)\rangle;\vt) + r(\vw)$. This includes SVM, logistic regression and all kinds of supervised learning as mentioned in the summary of \cite{stochastic}.} are included in the set of uniform convergence. These are the stared rectangle in Figure~\ref{stochastic.eps}.

At the same time, most of the problems (satisfying convexity, Lipschitz, boundedness, etc) are learnable (the triangle in Figure~\ref{stochastic.eps}), but not necessarily using empirical minimization, and not necessarily guaranteeing uniform convergence.
A generic way to learn such function is via online convex optimization. One may refer to Section \ref{sec:step1} for the algorithm, and its generalization guarantee is summarized in Eq.(7) and Eq.(8) of \cite{stochastic}.

Remark: All of the generalization error guarantees mentioned above are either in a factor of $\frac{1}{\sqrt{m}}$ or in a factor of $\frac{1}{m}$, depending on whether the objective function is convex of strongly-convex.

The main contribution of \cite{stochastic} is that they showed all Lipschitz-continuous strongly convex problems (dotted rectangle in Figure~\ref{stochastic.eps}) are learnable with empirical minimization.
This means that if one finds the empirical minimizer $\hat{\vw}$, $F(\hat{\vw})$ and $F(\vw^*)$ are guaranteed to be close enough.
Of course, there is also a good guarantee for any sub-optimal empirical minimizer $\vw$:
$$ F(\vw)-F(\vw^*) \leq \sqrt{\frac{2L^2}{\lambda}} \sqrt{\hat{F}(\vw)-\hat{F}(\hat{\vw})} + \frac{4L^2}{\delta\lambda m}\enspace,$$
here $L$ is the Lipschitz continuity constant for $f$, $\lambda$ is the strong-convexity constant for $f$, and $\delta$ is the confidence level.

\section{Trial 3: Stochastic Loss Optimization}\label{sec:step3}
Because the generalization bound differs between convex and strongly-convex objectives,
 we usually have to add a (strongly-convex) regularizer to ensure a better error bound between the empirical and stochastic objective.
For instance, we can add a $L_2$-norm regularizer to the hinge loss, resulting in the famous SVM problem.

If we have $f(\vw;\vt)=l(\vw;\vt)+r(\vw)$, we can define $F(\vw)=L(\vw)+r(\vw)$ where $L(\vw)=\mathbb{E}_{\vt \sim D}[l(\vw;\vt)]$. Then the mathematical term we are more interested is actually:
$$ L(\tilde{\vw}) - L(\vw_0) \enspace, $$
for a solution $\tilde{\vw}$ given by the algorithm, and the loss minimizer $\vw_0=\mathop{\textrm{argmin}}_{\vw} L(\vw)$.
By using the oracle inequality \cite{inverse,inverse2}:
\begin{eqnarray*}
&& L(\tilde{\vw})-L(\vw_0) \\
&=& (F(\tilde{\vw})-F(\vw^*)) + (F(\vw^*)-F(\vw_0)) \\
&-& r(\tilde{\vw}) + r(\vw_0) \\
&\leq& (F(\tilde{\vw})-F(\vw^*)) + r(\vw_0) \enspace,
\end{eqnarray*}
one may deduce a bound of the generalized loss error given a generalized error $F(\tilde{\vw})-F(\vw^*)$, while the latter is already obtained in Section~\ref{sec:step2}. Though this bound looks loose (neglecting two negative terms), through a careful selection of the weight hidden in the regularizer $r(\vw)$, one may find that the practical behavior matches this theoretical bound, in \cite{inverse,inverse2}.

\section{Trial 4: Zero-One Loss}\label{sec:step4}
Not satisfied by the result in the previous section, \cite{zeroone} made an interesting attempt towards learning 0-1 objective functions.
In classification problems with a half-plane classifier, the following objective function is more desirable than any other (regularized or not) convex objective (e.g. hinge loss, logistic loss):
$$f(\vw;\vt=(\vx,y))=\left| \frac{1}{2}\textrm{sgn}(\langle \vw, \phi(\vx)\rangle + \frac{1}{2} - y \right|$$
Notice that the label $y\in \{0,1\}$.

\subsection{The theory}
If we define $\varphi_{0-1}(a)=\frac{1}{2}(\textrm{sgn}(a)+1)$, the above objective function is characterized by the following concept class
$$H_{\varphi_{0-1}} = \{ \vx \rightarrow \varphi_{0-1}(\langle \vw, \phi(\vx) \rangle) \} \enspace,$$
and we are interested in optimizing the following stochastic objective:
\begin{equation}\label{eq:zeroone_obj}
F(h) = \mathbb{E}_{(x,y)\sim D} [| h(x)-y |], \quad h \in H_{\varphi_{0-1}} \enspace.
\end{equation}

The first step of this attempt requires the approximation to $H_{\varphi_{0-1}}$ using Lipschitz continuous functions.
Define
$$ \varphi_{\mathrm{sig}}(a) = \frac{1}{1+\exp(-4La)}\enspace, $$
which is L-Lipschitz continuous and approximates $\varphi_{0-1}$ well.\footnote{In \cite{zeroone} they also analyzed other two approximated functions, but for the lack of space they are ignored here.}
Now consider the following concept class for the stochastic objective (Eq.~\ref{eq:zeroone_obj})
$$ H_{\varphi_{\textrm{sig}}} = \{ \vx \rightarrow \varphi_{\textrm{sig}}(\langle \vw, \phi(\vx) \rangle) \} \enspace. $$

One advantage of such approximation is to allow theorems like Rademacher generalization bound \cite{Rademacher} to hold.
Indeed, the empirical minimizer,
$$ \hat{\vw} = \mathop{\textrm{argmin}}_{\vw} \hat{F}(\vw) = \mathop{\textrm{argmin}}_{\vw}  \frac{1}{m} \sum_{i=1}^m | \varphi_{\textrm{sig}}(\langle \vw, \vx_i \rangle) - y_i | \enspace, $$
gives a generalization error bound $F(\hat{\vw}) - F(\vw) < \epsilon$ when $m=\tilde{\Omega}(L^2/\epsilon^2)$.
However, it has been pointed out that the empirical minimization is ``hard'' since the objective is not convex.~\cite{zeroone}

To conquer such difficulty, a new concept class is introduced:
$$ H_B = \{ \vx \rightarrow \langle \vw, \psi(\vx)\rangle : \| \vw \|^2 \leq B \} \enspace, $$
and its difference from $H_{\varphi_{0-1}}$ or $H_{\varphi_{\textrm{sig}}}$ is twofold.
First, it no longer uses a 0-1 function in the prediction; the traditional half-plane classification using inner-product is adopted.
Second, it enables a new kernel $\psi$, which is defined as\footnote{$\nu$ can be chosen to be $1/2$ for the ease of presentation.}:
$$ \langle \psi(\vx), \psi(\vx') \rangle = \frac{1}{1-\nu \langle \phi(\vx), \phi(\vx') \rangle} \enspace. $$

Choosing $B=\Omega(\exp(L\log(\frac{L}{\epsilon})))$ large enough, \cite{zeroone} proved that $H_B$ approximately includes $H_{\varphi_{\textrm{sig}}}$,
 and therefore we can directly study the learning problem in $H_B$.
This only requires the Lipschitz continuity of $\varphi_{\textrm{sig}}$ and Chebyshev approximation technique, and is a very general proof.

One big benefit of such conversion is that the new problem is {\em convex} and can be empirically optimized via for instance stochastic gradient descent mentioned in Section~\ref{sec:step1}.
Pay attention that due to the boundedness of $H_B$, using Rademacher complexity bound again \cite{Rademacher, Kakade08}, a sample complexity of $\tilde{\Omega}(B/\epsilon^2)$ can be deduced.

The procedure above is {\em improper} learning: to learn $H_{\varphi_{0-1}}$ we actually incorporate a larger concept class $H_B$, and a classifier $h\in H_B$ will be returned which is close to the optimal classifier in $H_{\varphi_{0-1}}$.
Furthermore, the overall time and sample complexity is $\mathrm{poly}(\exp(L \log(\frac{L}{\epsilon})))$.
This bound is exponential w.r.t. $L$, but \cite{zeroone} also showed that a polynomial dependency on $L$ is impossible, unless some NP-hard problem is in $P$.

\subsection{The experiment}
Though the complexity depends crucially on $L$, however, when $L$ is a constant (e.g. $1$) there is still reason to believe that the approximated 0-1 loss function using $\varphi_{\mathrm{sig}}$ is more accurate than the hinge loss.
However, even if $L$ is constant, the sample complexity $m=\Omega(B/\epsilon^2)=\Omega(1/\epsilon^3)$ has cubic dependency on $\epsilon$.
At the same time, the kernel stochastic gradient descent algorithm (in the distribution manner) requires a time complexity of $O(m^2)$.
So there comes a dilemma: neither can we set $m$ to be too large since otherwise the empirical optimization cannot finish in endurable time, nor can we set $m$ to be too small since otherwise the generalization guarantee is not good enough.

Consider some $m$ of medium size, just enough to be trained in several minutes for example.
Though zero-one loss does not have a good generalization guarantee, however, it is a more desirable function than the hinge loss.
So given training set of medium size, and we \textbf{run SVM against a zero-one loss minimizer, who is to win this tug-of-war?}

\subsubsection{Configuration}
In the interest of fairness, the same stochastic gradient descent routine called P-packSVM~\cite{P-pack} has been adopted for the experiment, on an Intel Quad CPU machine with four cores (4 times speed-up).
Regarding the empirical optimization for the zero-one loss, two different approaches are implemented: \texttt{ZeroOne} refers to the classical mirror descent in Thm.~\ref{thm:mirror1}, and \texttt{ZeroOne-reg} refers to the regularized zero one loss objective with strongly-convex mirror descent in Thm.~\ref{thm:mirror2}.
At last, \texttt{PEGASOS} refers to the empirical optimizer for regularized SVM algorithm.

The three datasets {\em Splice}, {\em Web} and {\em Adult} presented by the libSVM project team \cite{libsvm} are used.
\begin{center}
\begin{tabular}{c|c|c|c}
   & \# Training / Testing Samples & \# Features \\
  \hline
  Splice & 1000 / 2175 & 60 \\
  Web & 2477 / 47272 & 300 \\
  Adult & 1605 / 30956 & 123 \\
  \hline
\end{tabular}
\end{center}

\subsubsection{Results}
\begin{table*}[ht!]
  \small
  \centering
  \caption{The accuracy report for three different methods on three different datasets, with the best-tuned parameters listed. The program has been run 5 times and the mean accuracy is chosen. The number of iterations is 100000 and all program runs in 30 seconds.}
    \begin{tabular}{l|ccc|ccc|ccc|}
    \toprule
          & \multicolumn{3}{c}{\texttt{PEGASOS}} & \multicolumn{3}{|c|}{\texttt{ZeroOne}} & \multicolumn{3}{c|}{\texttt{ZeroOne-reg}} \\
     & Accuracy & rbf   & $\lambda$     & Accuracy & rbf   & $\lambda$     & Accuracy & rbf   & $\lambda$ \\
    \midrule
    {Splice (Gaussian)} & 0.90069 & 0.02  & 0.0003 & \textbf{0.903448} & 0.01  & 0.08  & \textbf{0.902989} & 0.01  & 0.0006 \\
    {Splice (Linear)} & 0.846897 & -     & 0.0006 & \textbf{0.885517} & -     & 0.01  & \textbf{0.877701} & -     & 0.0003 \\
    {Adult (Gaussian)} & \textbf{0.844004} & 0.025 & 0.0003 & 0.84016 & 0.0125 & 0.003 & 0.840742 & 0.0125 & 0.0002 \\
    {Adult (Linear)} & \textbf{0.842971} & -     & 0.0003 & 0.838674 & -     & 0.02  & 0.838448 & -     & 0.0003 \\
    {Web (Gaussian)} & 0.980094 & 0.0125 & 0.00003 & \textbf{0.980729} & 0.0125 & 0.001 & \textbf{0.981236} & 0.0125 & 0.0001 \\
    {Web (Linear)} & 0.976667 & -     & 0.0003 & \textbf{0.981152} & -     & 0.003 & \textbf{0.980411} & -     & 0.0003 \\
    \bottomrule
    \end{tabular}
  \label{tab:accuracy}
\end{table*}
From Table~\ref{tab:accuracy}, one can see that when $m$ is on the magnitude of $1000$, it is still unclear that which method outperforms which.
For the dataset of Adult, the traditional SVM trainer \texttt{PEGASOS} significantly beats the zero one loss optimizers;
 while back to Splice and Web, zero-one loss does slightly better.

  \begin{figure*}[ht!]
    \centering
    \includegraphics[width=2\columnwidth]{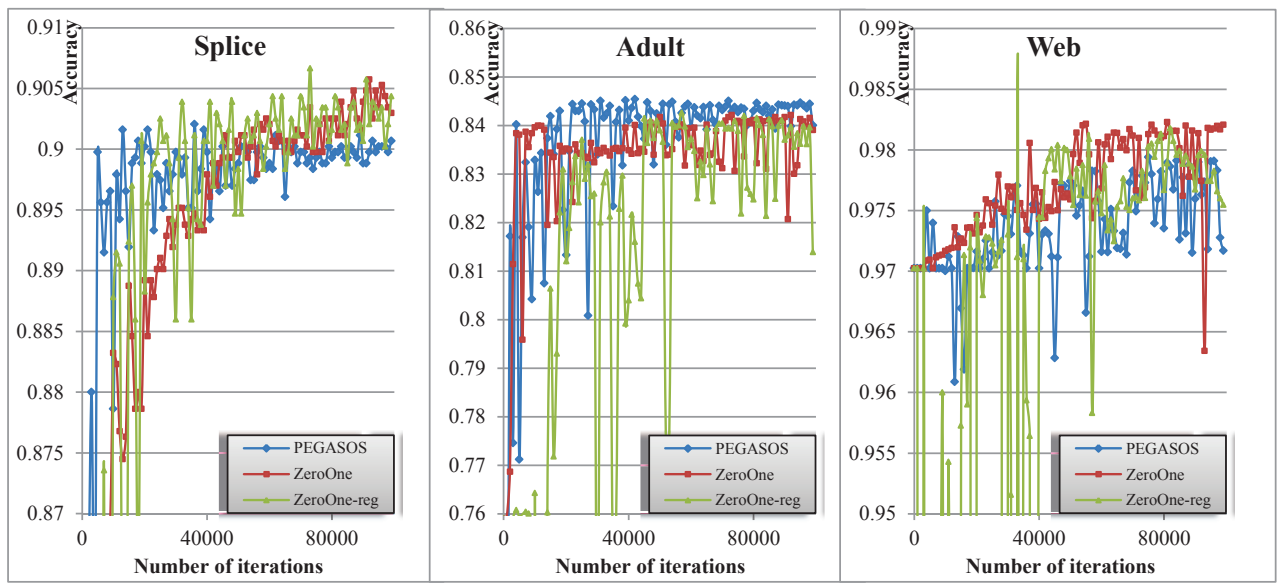}
    \caption{The variation of the accuracy as the number of training iterations increases.}
    \label{data.eps}
  \end{figure*}
Another experiment worth conducting is the convergence rate of the three methods.
From Figure~\ref{data.eps} one can see that because \texttt{ZeroOne} uses a learning rate of $1/\sqrt{T}$ where $T$ is the total number of iterations, the accuracy curve is smoother than that of \texttt{PEGASOS} and \texttt{ZeroOne-reg}, who uses learning rate $1/T$ for strongly-convexity.
However, because the generalization guarantee for 0-1 loss is weaker than SVM, \texttt{ZeroOne-reg} converges much slower than \text{PEGASOS}.

Notice that Web is a highly-biased dataset and 97 percent of the samples are negative. This explains that why all methods seem to perform with great turbulence on this dataset.

\bibliography{project} \bibliographystyle{alpha}

\newcommand{\etalchar}[1]{$^{#1}$}
\begin{thebibliography}{DSSST10}

\bibitem[BM03]{Rademacher}
Peter~L. Bartlett and Shahar Mendelson.
\newblock Rademacher and gaussian complexities: risk bounds and structural
  results.
\newblock {\em J. Mach. Learn. Res.}, 3:463--482, March 2003.

\bibitem[DSSST10]{Mirror}
John Duchi, Shai Shalev-Shwartz, Yoram Singer, and Ambuj Tewari.
\newblock Composite objective mirror descent.
\newblock In {\em COLT}, 2010.

\bibitem[Fan]{libsvm}
Rong-En Fan.
\newblock {\em {LIBSVM Data}: Classification, Regression, and Multi-label}.
\newblock Available at
  \url{http://www.csie.ntu.edu.tw/~cjlin/libsvmtools/datasets/}.

\bibitem[KST08]{Kakade08}
Sham~M. Kakade, Karthik Sridharan, and Ambuj Tewari.
\newblock On the complexity of linear prediction: Risk bounds, margin bounds,
  and regularization.
\newblock In {\em NIPS}, pages 793--800, 2008.

\bibitem[SSS08]{inverse}
Shai Shalev-Shwartz and Nathan Srebro.
\newblock Svm optimization: inverse dependence on training set size.
\newblock In {\em ICML}, pages 928--935, 2008.

\bibitem[SSSS07]{PEGASOS}
Shai Shalev-Shwartz, Yoram Singer, and Nathan Srebro.
\newblock Pegasos: Primal estimated sub-gradient solver for svm.
\newblock In {\em ICML}, pages 807--814, 2007.

\bibitem[SSSS10]{zeroone}
Shai Shalev-Shwartz, Ohad Shamir, and Karthik Sridharan.
\newblock Learning kernel-based halfspaces with the zero-one loss.
\newblock In {\em COLT}, 2010.

\bibitem[SSSSS09]{stochastic}
Shai Shalev-Shwartz, Ohad Shamir, Nathan Srebro, and Karthik Sridharan.
\newblock Stochastic convex optimization.
\newblock In {\em COLT}, 2009.

\bibitem[ZCW{\etalchar{+}}09]{P-pack}
Zeyuan~Allen Zhu, Weizhu Chen, Gang Wang, Chenguang Zhu, and Zheng Chen.
\newblock P-packsvm: Parallel primal gradient descent kernel svm.
\newblock In {\em ICDM}, pages 677--686, 2009.

\bibitem[ZCZ{\etalchar{+}}09]{inverse2}
Zeyuan~Allen Zhu, Weizhu Chen, Chenguang Zhu, Gang Wang, Haixun Wang, and Zheng
  Chen.
\newblock Inverse time dependency in convex regularized learning.
\newblock In {\em ICDM}, pages 667--676, 2009.

\end{thebibliography}

\end{document}